\documentclass[conference]{IEEEtran}
\IEEEoverridecommandlockouts
\usepackage{cite}
\usepackage{amsmath,amssymb,amsfonts}
\usepackage{algorithmic}
\usepackage[ruled,vlined]{algorithm2e}
\usepackage{graphicx}
\usepackage{textcomp}
\usepackage{xcolor}
\usepackage[shortlabels]{enumitem}
\newtheorem{remark}{Remark}
\def\BibTeX{{\rm B\kern-.05em{\sc i\kern-.025em b}\kern-.08em
    T\kern-.1667em\lower.7ex\hbox{E}\kern-.125emX}}
\begin{document}

\title{Energy-Efficient Lane Changes Planning and Control for Connected
Autonomous Vehicles on Urban Roads 
\thanks{$^{\dagger}$These authors equally contributed to this work.}
}

  

\author{\IEEEauthorblockN{Eunhyek Joa$^{\dagger 1}$ \qquad Hotae Lee$^{\dagger 1}$ \qquad Eric Yongkeun Choi$^{1}$ \qquad Francesco Borrelli$^{1}$} 
\IEEEauthorblockA{$^{1}$Mechanical Engineering, University of California, Berkeley, USA}}

\maketitle

\begin{abstract}
This paper presents a novel energy-efficient motion planning algorithm for Connected Automated Vehicles (CAVs) on urban roads. 
The approach consists of two components: a decision-making algorithm and an optimization-based trajectory planner. 
The decision-making algorithm leverages Signal Phase and Timing (SPaT) information from connected traffic lights to select a lane with the aim of reducing energy consumption.
The algorithm is based on a heuristic rule which is learned from human driving data. 
The optimization-based trajectory planner generates a safe, smooth, and energy-efficient trajectory toward the selected lane.
The proposed strategy is experimentally evaluated in a Vehicle-in-the-Loop (VIL) setting, where a real test vehicle receives SPaT information from both actual and virtual traffic lights and autonomously drives on a testing site, while the surrounding vehicles are simulated. 
The results demonstrate that the use of SPaT information in autonomous driving leads to improved energy efficiency, with the proposed strategy saving 37.1\% energy consumption compared to a lane-keeping algorithm.
\end{abstract}


\section{Introduction}
Studies on Connected and Automated Vehicles (CAVs) have gained substantial interest in the automotive industry due to their potential to improve road safety, increase energy efficiency, and optimize road utilization \cite{guanetti2018CAVs}.
CAV technology implementation is enabled by utilizing advanced connectivity solutions such as vehicle-to-infrastructure (V2I), vehicle-to-vehicle (V2V), and vehicle-to-cloud (V2C) communications, facilitating coordination and collaboration between traffic elements like traffic signals and surrounding vehicles.
Cooperative Adaptive Cruise Control (CACC) systems, for instance, leverages V2V communications to sustain string stability even at close inter-vehicle distances \cite{naus2010string}, resulting in significant improvement of energy efficiency in highway driving scenarios \cite{mcauliffe2018CACC, kim2021CACC}.
Also, it has been shown that CACC systems enhance road utilization \cite{askari2017CACCthroughput} as well as energy performance \cite{bertoni2017CACCurban} in urban driving scenarios.

In urban driving contexts there has been extensive research on energy-efficient CAV technology utilizing V2I communications at signalized intersections.
The concept is that Signal Phase and Timing (SPaT) information is transmitted to surrounding vehicles and used for improved driving comfort and energy conservation through reduced stops \cite{sciarretta2020CAVs}. 
For instance, \cite{bae2019real} presents a two-level receding horizon control framework utilizing real-time and historical SPaT information from multiple traffic lights for a single vehicle. 
Similarly, \cite{ard2023VILCAV} develops a real-time capable eco-driving controller using Pontryagin's Minimum Principle approach for a single vehicle. 
The work in \cite{wang2019V2IV2V} employs both V2V and V2I communications to create cooperative eco-driving, reducing energy consumption and pollutant emissions for multiple CAVs.
In the aforementioned work and in the majority of existing literature on energy-efficient CAVs, the focus has been on energy-efficient longitudinal control of single-lane vehicles. \emph{This paper focused  on energy efficient lateral control}.
Validating CAV technology in real-world settings is desirable, however, the complexities and variability of real-world traffic and safety concerns can pose challenges \cite{ma2018VIL}. 
In order to mitigate these challenges associated with evaluating the performance of the proposed control system, a Vehicle-in-the-Loop (VIL) setup utilizing a microscopic simulator is utilized in \cite{bae2019VIL, ard2021energyVIL}.
This allows us to test the performance of the proposed control system in a more realistic yet safe environment, while still leveraging the benefits of simulation for perception and prediction.
In \cite{bae2019VIL}, before being tested on the real environment in \cite{bae2022ecological}, the ecological adaptive cruise controller was evaluated in simulated traffic while the test vehicle was put on top of a chassis dynamometer.
A VIL setup was designed in \cite{ard2021energyVIL} to evaluate cruise controllers for CAVs. The setup enables autonomous driving on a physical testing site while interacting with virtual surrounding vehicles in a simulated environment.
Due to safety and legal considerations, testing algorithms for controlling vehicles in both longitudinal and lateral directions on real urban roads is challenging. To overcome this, in this paper, a VIL setup is constructed to test the proposed algorithm on a closed-track environment.

We present a novel energy-efficient motion planning strategy for CAVs that exploit the benefits of opportunistic lane changes in urban driving scenarios with the presence of traffic lights.
The paper contribution is twofold:
\begin{itemize}
    \item A novel energy-efficient motion planning algorithm to exploit the benefits of lane changes by leveraging V2I communications.
    \item Experimentally demonstrating energy savings in a VIL setting with a hybrid vehicle where total energy consumption is calculated from measurements of accurate fuel flow sensor and voltage/current meters.
\end{itemize}
The remainder of the paper is organized as follows. Section II details the proposed control architecture. Sections III and IV explain the proposed energy-efficient planning algorithm. Section V provides details on the designed VIL setup and experiment results. Finally, in Section VI, the paper concludes with future work.

\section{Control Architecture}
\begin{figure}[h!]
    \centering
    \vspace{-1.0em}
    \includegraphics[width=0.90\columnwidth]{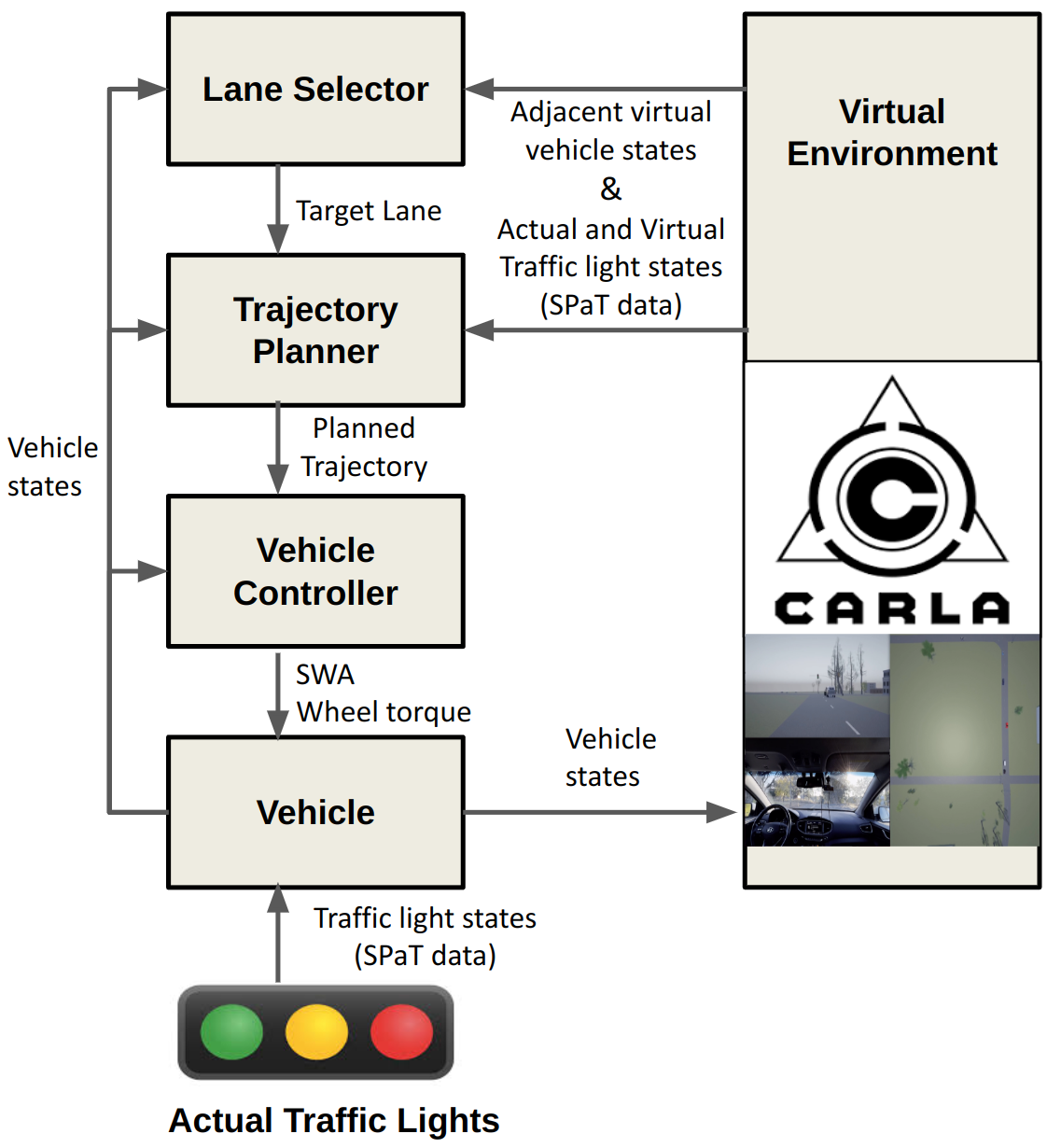}
    \caption{Diagram of Control Architecture}
    \vspace{-1.0em}
    \label{fig:control_architecture}
\end{figure}

The overall block diagram of control architecture and the entire system is illustrated in Fig. \ref{fig:control_architecture}.
The developed hierarchical control system consists of a lane selector, a trajectory planner, and a vehicle controller. 
The lane selector determines the target lane for the ego vehicle. 
The trajectory planner employs optimization-based trajectory generation for lane keeping and lane change maneuvers. 
The energy consumption minimization problem with safety guarantee is solved for the lane-keeping maneuver. The minimization problem of the lateral error to a target lane and the discomfort of passengers is solved for the lane change maneuver.
The vehicle controller is comprised of a tracking controller utilizing a Model Predictive Controller (MPC) to ensure accurate tracking of the planned trajectory and an actuator controller which employs a combination of classical feedback control with feedforward inputs and adaptive control for wheel torque and steering angle. 
The following sections describe the details of the lane selector and the trajectory planner in the control architecture.
For the details of the tracking controller and actuator controller, see \cite{kim2023trajectory}, \cite{adaptive_control_steering_control}. 

The virtual environment synchronizes the ego vehicle and physical traffic light in the testing site by receiving the actual vehicle states and SPaT information from the real-world counterparts.
The virtual environment simulates all virtual agents that interact with the ego vehicle as its global position and speed are known.
The virtual environment returns updates on the vehicle states of surrounding vehicles and SPaT information of both actual and virtual traffic lights to the integrated controller.

 
\section{Lane Selector}
To minimize energy consumption while driving in heavy traffic, it is crucial to decide which lane the ego vehicle moves on. 
The proposed strategy is derived from an analysis of multiple human-driven data sets. 
Driving data was collected from multiple participants to extract the fundamental principle for evaluating the most energy-efficient lane in terms of future energy consumption.
Based on our observations, a reduction in the number of full stops significantly impacts energy consumption. Therefore, our strategy targets minimizing the number of full stops by choosing an appropriate lane to pass the traffic light using SPaT information. 

The lane selector determines the target lane for the ego vehicle and whether the ego vehicle tries to pass the current traffic light or not in the target lane using SPaT information. 
If the selected lane is different from the current lane, the lane selector sends a signal indicating a lane change opportunity, prompting the ego vehicle to initiate a lane change maneuver if available.

\begin{figure}[ht]
    \centering    \includegraphics[width=0.75\columnwidth]{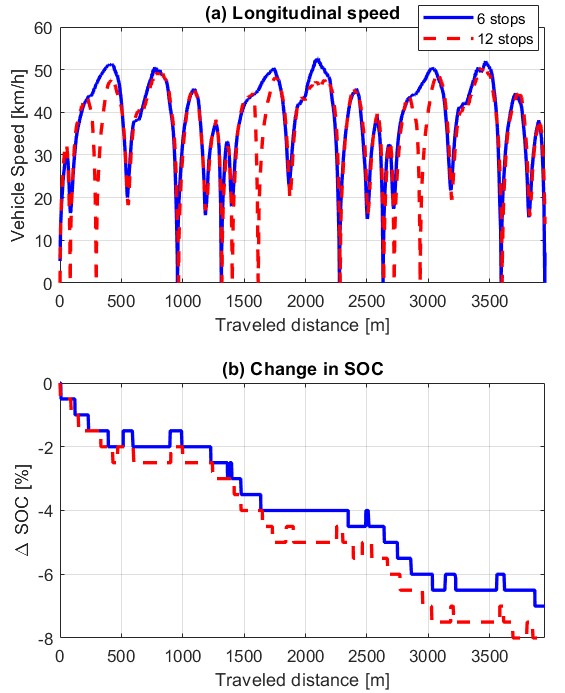}
    \caption{Human driving data: Two tests with different numbers of stops}
    \vspace{-1.0em}
    \label{fig:human-driving}
\end{figure}

Examples of collected human driving trajectories are shown in Fig. \ref{fig:human-driving}.
A similar speed profile is maintained across the two tests while the number of stops changed: one test shows six stops (depicted as a blue solid line), while the other test shows 13 stops (depicted as a red dashed line).
Though the maximum and the average speed of the blue line are higher than that of the red dashed line, the total energy consumption of the blue line is 26.4\% lower than that of the red dashed line.
Companies with abundant availability of human driving data can use the same approach and learn a lane change strategy from a large real-world dataset.

\subsection{Strategy}
This paper focuses on a two lane, one-way road, however, the methodology presented can be extrapolated to encompass multiple lane roads.
The lane selector evaluates the feasibility of passing the upcoming traffic light for each lane.
The algorithm utilizes various data inputs including the current phase and remaining duration of the nearest traffic light, as well as the speed and position information of both the ego vehicle and any preceding vehicles in each lane under consideration. 
It is assumed that the data can be obtained from connectivity and perception. 
Once the assessments of each lane in terms of passing the traffic light are finished, the lane selector makes a final determination of the target lane for the ego vehicle and sends a signal indicating a lane change opportunity to the trajectory planner.
The strategy is based on the assumption that the traffic light signal is deterministic.
\subsubsection{Notation}
For brevity of explanation, let $\mathcal{C}_{\mathrm{pass}}$ denote the decision of whether the ego vehicle attempts to pass the nearest traffic light or not, i.e., $\mathcal{C}_{\mathrm{pass}} \in \{\text{PASS}, \text{NONPASS}\}$. $ s_{}, v_{}, s^{\mathrm{front}}, v^{\mathrm{front}}, s_{\mathrm{tl}}, T_{\mathrm{tl}}, p_{\mathrm{tl}}$ denote a longitudinal distance of the ego vehicle, a longitudinal speed of the ego vehicle, a longitudinal distance of the preceding vehicle, a longitudinal speed of the preceding vehicle, a longitudinal distance of the current traffic light, the remaining time of the current traffic light and the current phase of the traffic light, respectively. 
\subsubsection{Rule to assign a value to $\mathcal{C}_{\mathrm{pass}}$}
In this section, we present a succinct guideline for determining $\mathcal{C}_{\mathrm{pass}}$ depending on the current phase of the traffic light. In the determination of $\mathcal{C}_{\mathrm{pass}}$, the main factors are the remaining time of the current traffic light phase and the estimated time it would take for the ego vehicle to reach the traffic light. When $p_{\mathrm{tl}}$ is `GREEN', if $T_{\mathrm{tl}}$ is greater than the estimated time to reach the traffic light, $\mathcal{C}_{\mathrm{pass}}$ is determined as `PASS'. Otherwise, $\mathcal{C}_{\mathrm{pass}}$ is determined as `NONPASS'.
On the contrary when $p_{\mathrm{tl}}$ is `RED', if $T_{\mathrm{tl}}$ is greater than the estimated time to reach the traffic light, $\mathcal{C}_{\mathrm{pass}}$ is determined as `NONPASS'. If $T_{\mathrm{tl}}$ is smaller, the ego vehicle can pass the traffic light at the next `GREEN' phase without a full stop. Therefore, $\mathcal{C}_{\mathrm{pass}}$ is determined as `PASS'. When $p_{\mathrm{tl}}$ is `YELLOW', we decide $\mathcal{C}_{\mathrm{pass}}$ as `NONPASS' conservatively. Since the `YELLOW' phase does not remain for a long time, the lane selector shortly updates the determination of $\mathcal{C}_{\mathrm{pass}}$  based on the case of the `RED' phase.  

The estimated time computation for the ego vehicle to reach the traffic light is approximated based on the following assumptions.
\begin{enumerate}[i)]
    \item The preceding vehicle maintains a constant speed.
    \item The ego vehicle maintains a constant speed unless the preceding vehicle is slower than the ego vehicle. If the preceding vehicle is slower than the ego vehicle, the ego vehicle maintains a current constant speed until it reaches the preceding vehicle. Upon reaching the preceding vehicle, the ego vehicle reduces its speed to match that of the preceding vehicle. 
\end{enumerate}
For example, the time required for the ego vehicle to catch up with the preceding vehicle traveling at a slower speed can be calculated as $\frac{s^{\mathrm{front}}-s}{v-v^{\mathrm{front}}}$. If the ego vehicle is yet to reach the traffic light at this point, the estimated time for the ego vehicle to reach the traffic light after catching up the front vehicle is calculated as $\frac{s_{\mathrm{tl}}(v-v^{\mathrm{front}})-v(s^{\mathrm{front}}-s)}{(v-v^{\mathrm{front}})v^{\mathrm{front}}}$. The sum of them is the estimated time for ego vehicle to reach the traffic light in this example.

\subsubsection{Rule to assign the target lane}
Once the determination of $\mathcal{C}_{\mathrm{pass}}$ is done for each lane in the previous step, the lane selector makes a final decision of the target lane based on the resulting $\mathcal{C}_{\mathrm{pass}}$. 
If $\mathcal{C}_{\mathrm{pass}}$ is `PASS' for only one lane, the target lane is the lane in which the ego vehicle is able to pass the traffic light. 
If $\mathcal{C}_{\mathrm{pass}}$ for all lanes are `NONPASS', the ego vehicle keeps the current lane (i.e. the target lane is the current lane). 
If $\mathcal{C}_{\mathrm{pass}}$ for all lanes are `PASS', the lane selector chooses the lane that requires the least amount of time to reach the traffic light, in an effort to increase the likelihood of passing subsequent traffic lights.

\section{Optimization-based Trajectory Planner}
The trajectory planner, operating at a frequency of 1Hz, generates trajectories that are smooth for passengers' comfort and compliant with vehicle dynamics constraints and safety constraints.
Specifically, the planned trajectory is composed of 50 waypoints with a time interval of 0.1 sec between each point, resulting in a total prediction horizon of 5 sec. Each waypoint is associated with calculated values for heading angle, curvature, speed, and longitudinal acceleration.

We categorize urban road driving scenarios into \textit{Lane Keeping} and \textit{Lane Change} cases.
The \textit{Lane Keeping} category includes all driving scenarios where no lane change is required, such as taking a left or right turn, maintaining a safe distance from the front vehicle, and stopping at a stop sign. 
The \textit{Lane Change} category includes scenarios where lane changes are required. The main difference between the two categories is that in a \textit{Lane Change} scenario, a lateral motion must be planned to safely move into the target lane, while in a \textit{Lane Keeping} scenario, the vehicle must simply stay within the current lane. 
To address these differing requirements, we have designed two separate planners: one for \textit{Lane Keeping} scenarios and one for \textit{Lane Change} scenarios. 
The appropriate planner is selected based on a lane change indicator, which is determined by the lane selector.

\begin{figure}[b!]
    \centering
    \vspace{-1.0em}
    \includegraphics[width=0.6\columnwidth]{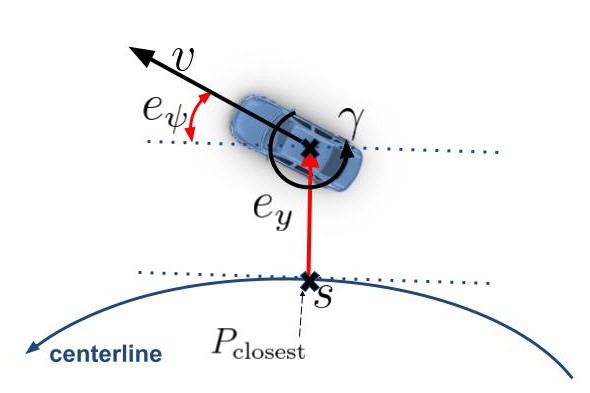}
    \vspace{-1.0em}
    \caption{Point mass model for planner}
    \vspace{-1.0em}
    \label{fig:point mass model}
\end{figure}
\subsection{Vehicle Model}
The point mass model illustrated in Fig. \ref{fig:point mass model} is used to design the trajectory planners.
Note that as the operating design domain of our system is limited to low-speed urban roads, we assume that lateral speed is negligible.

The closest point $P_\mathrm{closest}$ is identified and the system states are described with respect to this point.
$s$ is the traveled distance of $P_\mathrm{closest}$ along the centerline, $v$ is the vehicle speed, $e_y$ is the lateral deviation from the $P_\mathrm{closest}$, and $e_\psi$ is the heading error.
We also use the curvature of the centerline at the point $P_\mathrm{closest}$, which we denote as $\kappa_\mathrm{road}$.
The system equation can be described as:
\begin{equation} \label{eq: point mass model}
    \begin{split}
        & \mathbf{x} = \begin{bmatrix} s & v & e_y & e_\psi \end{bmatrix}^\top, \quad \mathbf{u} = \begin{bmatrix} T_\mathrm{whl} & \kappa \end{bmatrix}^\top \\
        & \Dot{\mathbf{x}} = \begin{bmatrix} v \cos{e_\psi} \\ 0 \\ v \sin{e_\psi} \\ - \kappa_\mathrm{road} v \cos{e_\psi} \end{bmatrix} + \begin{bmatrix} 0 & 0 \\ \frac{1}{m r_\mathrm{eff}} & 0 \\  0 & 0 \\ 0 & v \cos{e_\psi} \end{bmatrix} \mathbf{u}\\
    \end{split}
\end{equation}
where $\kappa$ is the curvature of the vehicle, $T_\mathrm{whl}$ is the total wheel torque, $r_\mathrm{eff}$ is the effective wheel radius, and $m$ is the vehicle mass.
We select the curvature $\kappa$ as the control input of the lateral motion because it is empirically known that the smoothness of the trajectory and the comfort while driving through the trajectory are related to the curvature \cite{gonzalez2014continuous}. 
We discretize the continuous-time model \eqref{eq: point mass model} using the forward Euler method as $\mathbf{x}_{k+1} = \mathbf{f}_d (\mathbf{x}_{k}) + \mathbf{g}_d (\mathbf{x}_{k}) \mathbf{u}_{k}$, where $\mathbf{x}_{k}$ and $\mathbf{u}_{k}$ are the state and the input at the time step $k$, respectively. The discretization time is 0.1 sec.

\subsection{Energy Consumption Model}
To minimize the energy consumption of the trajectory, we need to model an energy consumption stage cost function $\ell_\mathrm{e}(\mathbf{x}, \mathbf{u})$, which maps current vehicle states and inputs to the amount of energy consumption.
We utilize the following parametric model for the $\ell_\mathrm{e}(\mathbf{x}, \mathbf{u})$ based on \cite{sciarretta2015optimal}:
\begin{equation} \label{eq: energy consumption stage cost}
    \ell_\mathrm{e}(\mathbf{x}, \mathbf{u}) = c_1 T_\mathrm{whl} v + c_2 v 
\end{equation}
where $c_1$ and $c_2$ are parameters to be identified.
\begin{remark}
In this paper, as the test vehicle is a PHEV, we require two energy-consumption models - one for the combustion engine and another for the electric motor. 
Ideally, the variables for the stage cost would include an integer variable that indicates the current power source, such as battery-powered, engine-powered, or a combination of both so that we can optimize the power source.
However, in our research, we do not have the ability to choose a power source as it is determined by the manufacturer's logic. Therefore, instead of modeling the stage cost separately for each power source, we aim to find an average energy consumption model.
A better model could be used in \cite{Choi2021Thesis}.

\end{remark}

The parameters of \eqref{eq: energy consumption stage cost} are identified by solving the following regression problem with the pre-recorded dataset:
\begin{equation} \label{eq: regression problem}
\begin{split}
    & \min_{\substack{c_{1},c_2}} \sum_{i=1}^{N_\mathrm{data}} (c_1 T_\mathrm{whl, i} v_{i} + c_2 v_{i} - P_\mathrm{tot, i})^2 \\
    & \,\,\, \textnormal{s.t.,} \quad c_1 \geq 0, \,\, c_2 \geq 0,
\end{split}
\end{equation}
where a variable with the subscript $i$ represents $i$-th data point in the dataset, and the total power consumption, $P_\mathrm{tot, i}$, is calculated from measurements of a fuel flow sensor and battery sensors \footnote{The measured fuel flow is converted into the fuel power by multiplying the conversion factor. The battery sensors consist of voltage and current meters so the battery power is calculated by multiplying measured voltage and current.}.
The optimal values are $c_1^\star = 4.47$ and $c_2^\star = 1522.23$. 

\begin{figure}[h!]
    \centering
    \vspace{-1.0em}
    \includegraphics[width=0.95\columnwidth]{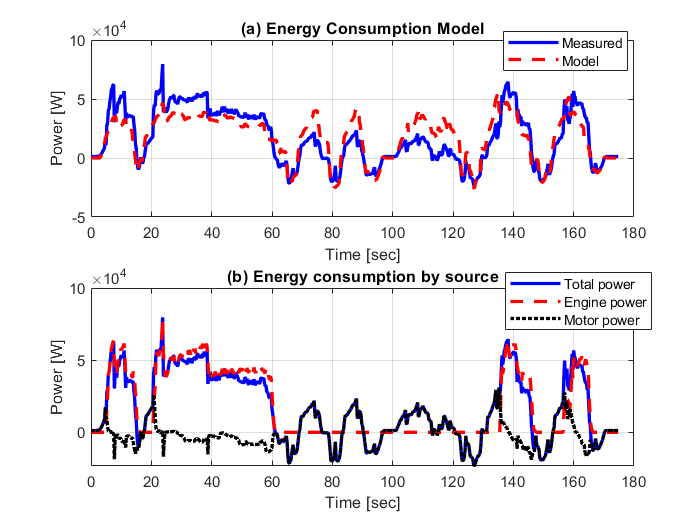}
    \vspace{-1.0em}
    \caption{Analysis and Results of the Energy model regression}
    \label{fig:energy regression result}
\end{figure}

We validated the fidelity of the energy consumption model using other datasets. 
One result is presented in Fig. \ref{fig:energy regression result}. 
As depicted in Fig. \ref{fig:energy regression result}, a discrepancy between the model's and the actual system's energy consumption is observed. 
This discrepancy can be attributed to the fact that we have used an average energy consumption model.
\begin{remark}
    Due to the regenerative braking, the stage cost in \eqref{eq: energy consumption stage cost} can be negative. However, including this cost without modification in the planner makes the vehicle remain at standstill. This is because the system thinks that it can gain energy if it decelerates and our speed sensor only gives the magnitude of speed, not the sign. Thus, to make our system work, we set our energy consumption stage cost as:
    \begin{equation} \label{eq: modified energy stage cost}
        \Bar{\ell}_\mathrm{e}(\cdot, \cdot) = \max\{\ell_\mathrm{e}(\cdot, \cdot),0\}
    \end{equation}
    Modifying the stage cost to include regenerative braking is the scope of our future research.
\end{remark}
\subsection{Lane Keeping}
The goal of the trajectory planner for lane keeping is to generate a smooth, energy-efficient speed trajectory while obeying the flow speed.
Regarding the corresponding lateral motion, we use the centerline of the route for the planned path, i.e., the planned curvature $\kappa = \kappa_\mathrm{road}$ and $e_y=e_\psi =0$.
Thus, we simplify the model in \eqref{eq: point mass model} as:
\begin{equation} \label{eq: point mass model for LK}
    \begin{split}
        & \mathbf{x}^\mathrm{LK} = \begin{bmatrix} s & v \end{bmatrix}^\top, \quad \mathbf{u}^\mathrm{LK} = T_\mathrm{whl} \\
        & \Dot{\mathbf{x}}^\mathrm{LK} = \begin{bmatrix} 0 & 1 \\ 0 & 0 \end{bmatrix} \mathbf{x}^\mathrm{LK} + \begin{bmatrix} 0 \\ \frac{1}{m r_\mathrm{eff}} \end{bmatrix} \mathbf{u}^\mathrm{LK}\\
    \end{split}
\end{equation}
We discretize this simplified model as $\mathbf{x}^\mathrm{LK}_{k+1} = A_d \mathbf{x}^\mathrm{LK}_k + B_d \mathbf{u}^\mathrm{LK}_k$. 
The discretization time is 0.1 sec.

\subsubsection{Cost function design}
We include three different aspects in our planner for lane keeping.
First, to minimize the energy consumption along the planned trajectory, we consider the sum of the energy consumption stage cost $\Bar{\ell}_\mathrm{e}(\cdot, \cdot)$ in \eqref{eq: modified energy stage cost}:
\begin{equation} \label{eq: lk energy cost}
    \begin{split}
    & J_\mathrm{energy}^\mathrm{LK}(\{\mathbf{x}^\mathrm{LK}_{i|k}\}_{i=0}^{N-1}, \{\mathbf{u}^\mathrm{LK}_{i|k}\}_{i=0}^{N-1}) = \sum_{i=0}^{N-1} \Bar{\ell}_\mathrm{e}(\mathbf{x}^\mathrm{LK}_{i|k}, \mathbf{u}^\mathrm{LK}_{i|k})    
    \end{split}
\end{equation}
where $\mathbf{x}^\mathrm{LK}_{i|k}$ and $\mathbf{u}^\mathrm{LK}_{i|k}$ are state and control input for predicted time step $k+i$ at current time step $k$, respectively.

Second, to generate a smooth speed trajectory, we consider a smoothness stage cost that consists of the first and second-order time derivatives of the planned speed. 
The first derivative of speed represents acceleration, while the second derivative represents jerk.
By minimizing the sum of this stage cost, we aim to minimize the magnitude of acceleration and jerk, as well as the duration of high jerk for passengers' comfort \cite{de2023standards}. 
\begin{equation} \label{eq: lk smooth cost}
    \begin{split}
    & J_\mathrm{smooth}^\mathrm{LK}(\{\mathbf{x}^\mathrm{LK}_{i|k}\}_{i=0}^{N-1}) = \sum_{i=0}^{N-2} (v_{i|k} - 2 v_{i+1|k} + v_{i+2|k})^2 \\
    & \qquad \qquad \qquad \qquad \quad  + \sum_{i=0}^{N-1} (v_{i|k} - v_{i+1|k})^2 
    \end{split}
\end{equation}

Third, to make our system drive with the flow of the traffic, we penalize the deviation of the current speed from the flow speed of the current lane.
\begin{equation} \label{eq: lk tracking cost}
    \begin{split}
    & J^\mathrm{LK}_\mathrm{tracking}(\{\mathbf{x}^\mathrm{LK}_{i|k}\}_{i=0}^{N-1}) = \sum_{i=0}^{N-1} (v_{i|k} - v_{\mathrm{ref},i|k})^2 
    \end{split}
\end{equation}
where the $v_{\mathrm{ref},i|k}$ is the predicted flow speed at time step $k+i$. $v_{\mathrm{ref},i|k}$ can be a legal speed or a front vehicle's speed.
The legal speed can be the maximum speed that is allowed legally or zero to stop before the stop sign.

\subsubsection{Constraints}
To ensure the safety of our system, we impose constraints on the states and the control inputs.
\begin{subequations}
\begin{align}
    & 0 \leq v \leq v_{\mathrm{max}}, \label{eq: LK state constraint} \\
    & T_\mathrm{brake} \leq T_\mathrm{whl} \leq T_\mathrm{motor}, \label{eq: LK input constraint} \\ 
    & ({s}^{\mathrm{front}} - s) \geq d_\mathrm{safe} + (v - {v}^{\mathrm{front}}) t_\mathrm{gap}, \label{eq: npc safety constraint}
\end{align}
\end{subequations}
where $v_{\mathrm{max}}$ is the maximum speed that is allowed legally, $T_\mathrm{brake}$ and $T_\mathrm{motor}$ are maximum torques for braking and traction, respectively, $d_\mathrm{safe}$ is a safe distance which is set to 10m, $t_\mathrm{gap}$ is a time gap, and ${s}^{\mathrm{front}}$ and ${v}^{\mathrm{front}}$ are the states of the front vehicle.
Note that \eqref{eq: npc safety constraint} is a collision avoidance with the front vehicle with constant speed motion assumption. 
To sum up, the feasible set $\mathcal{F}^\mathrm{LK}({s}^{\mathrm{front}}, {v}^{\mathrm{front}}) =\{(\mathbf{x}, \mathbf{u}): \eqref{eq: LK state constraint}, \eqref{eq: LK input constraint}, \eqref{eq: npc safety constraint}\}$

\subsubsection{Optimal Control Problem}
To sum up, we solve the following constrained optimal control problem for every step in receding horizon fashion:
\begin{equation} \label{eq:lane keeping planner}
\begin{split}
    & V^\mathrm{LK}_{k \rightarrow k+N} (\mathbf{x}^\mathrm{LK}_k, \mathbf{u}^\mathrm{LK}_k) = \\
    & \min_{\{\mathbf{u}^{\mathrm{LK}}_{i|k}\}_{i=0}^{N-1}} J_\mathrm{energy}^\mathrm{LK}(\{\mathbf{x}^\mathrm{LK}_{i|k}\}_{i=0}^{N-1}, \{\mathbf{u}^\mathrm{LK}_{i|k}\}_{i=0}^{N-1}) \\
    & \quad \quad \quad \,\, + J_\mathrm{smooth}^\mathrm{LK}(\{\mathbf{x}^\mathrm{LK}_{i|k}\}_{i=0}^{N-1})  + J^\mathrm{LK}_\mathrm{tracking}(\{\mathbf{x}^\mathrm{LK}_{i|k}\}_{i=0}^{N-1}) \\
    & \qquad \textnormal{s.t.,} \quad \quad \mathbf{x}^\mathrm{LK}_{0|k} = \mathbf{x}^\mathrm{LK}_k, \,\, \mathbf{u}^\mathrm{LK}_{0|k} = \mathbf{u}^\mathrm{LK}_k, \\
    & \quad \qquad \qquad \, \, \mathbf{x}^\mathrm{LK}_{i+1|k} = A_d \mathbf{x}^\mathrm{LK}_{i|k} + B_d \mathbf{u}^\mathrm{LK}_{i|k}, \\
    & \quad \qquad \qquad \, \, (\mathbf{x}^\mathrm{LK}_{i|k}, \mathbf{u}^\mathrm{LK}_{i|k}) \in \mathcal{F}^\mathrm{LK}(s^{\mathrm{front}}_{i|k}, v^{\mathrm{front}}_{i|k}),\\
    & \quad \qquad \qquad \, \, i=0,1,...,N-1,
\end{split}
\end{equation}
where $s^{\mathrm{front}}_{i|k}$ and $v^{\mathrm{front}}_{i|k}$ are the traveled distance and speed of the front vehicle at predicted time step $k+i$, respectively, with the assumption of the constant speed motion of the front vehicle. 
After solving \eqref{eq:lane keeping planner}, the planner transmits whole optimal sequences of state $\{\mathbf{x}^{\mathrm{LK},\star}_{i|k}\}_{i=0}^{N}$ and input $\{\mathbf{u}^{\mathrm{LK},\star}_{i|k}\}_{i=0}^{N-1}$ to the tracking controller.

\subsection{Lane Change}
The goal of the trajectory planner for lane change is to generate a smooth, safe trajectory to the target lane.

\subsubsection{Cost function design} For smooth trajectory, the first and second derivatives of speed and curvature are considered similar to the \eqref{eq: lk smooth cost}.
\begin{equation}
    \begin{split}
        & J_\mathrm{smooth}(\{\mathbf{x}_{i|k}\}_{i=0}^{N-1}, \{\mathbf{u}_{i|k}\}_{i=0}^{N-1}) = \\
        & \sum_{i=0}^{N-2} (v_{i|k} - 2 v_{i+1|k} + v_{i+2|k})^2 + \sum_{i=0}^{N-1} (v_{i|k} - v_{i+1|k})^2 +\\
        & \sum_{i=0}^{N-2} \rho_{\kappa 2} (\kappa_{i|k} - 2 \kappa_{i+1|k} + \kappa_{i+2|k})^2 + \sum_{i=0}^{N-1} \rho_{\kappa 1} (\kappa_{i|k} - \kappa_{i+1|k})^2 
    \end{split}
\end{equation}
We penalize the deviation of the terminal state from the target lane as follows:
\begin{equation}
    \begin{split}
        & J_\mathrm{target} (\mathbf{x}_{N|k}) = \rho_y (e_{y, N|k} - y_\mathrm{target})^2 + \rho_{\psi} e_{\psi,N|k}^2 
    \end{split}
\end{equation}
where $y_\mathrm{target}$ is the lateral value of the target lane with respect to the centerline, and $\rho_y$ and $\rho_{\psi}$ are the weighting factors.
In summary, the following cost is minimized:
\begin{equation}
    \begin{split}
        & J^\mathrm{LC}(\{\mathbf{x}_{i|k}\}_{i=0}^{N}, \{\mathbf{u}_{i|k}\}_{i=0}^{N-1}) = \\
        & J_\mathrm{smooth}(\{\mathbf{x}_{i|k}\}_{i=0}^{N-1}, \{\mathbf{u}_{i|k}\}_{i=0}^{N-1}) + J_\mathrm{target} (\mathbf{x}_{N|k}) 
    \end{split}
\end{equation}

\subsubsection{Constraints}
Besides \eqref{eq: LK state constraint} and \eqref{eq: LK input constraint}, we consider the following constraints for state and input:
\begin{equation} \label{eq: LC state input ay constraint}
    \begin{split}
        & e_{y, \mathrm{min}} \leq e_y \leq e_{y, \mathrm{max}}, \,\, -e_{\psi, \mathrm{bnd}} \leq e_\psi \leq e_{\psi, \mathrm{bnd}}, \\ 
        & -\kappa_\mathrm{bnd} \leq \kappa \leq \kappa_\mathrm{bnd}, \,\,  -a_{y,\mathrm{bnd}} \leq v^2 \kappa \leq a_{y,\mathrm{bnd}},\\
    \end{split}
\end{equation}
where $e_{y, \mathrm{max}}$ and $e_{y, \mathrm{min}}$ are the lateral values of road boundaries with respect to the centerline, $e_{\psi, \mathrm{bnd}}$ is set to $0.5\pi$, $\kappa_{\mathrm{bnd}}$ is set to 0.1, and $a_{y,\mathrm{bnd}}$ is set to $3m/s^2$.
The lateral acceleration bound is set as severe lateral acceleration affects the discomfort of passengers.

To safely change to the target lane, collision avoidance constraints in \cite{obca} are considered.
For real-time computation, we use the simple point-mass model formulation in \cite{obca} by setting the ego vehicle as the point mass and the surrounding vehicles as the enlarged polytope.
The collision avoidance constraints can be written as:
\begin{equation} \label{eq:obca constraint}
\begin{split}
    & (A^{(m)} \mathbf{x} -b^{(m)}(x^{\mathrm{sv},(m)}))^\top \lambda^{(m)} \geq d_\mathrm{min},\\
    & \lVert {A^{(m)}}^\top \lambda^{(m)} \rVert \leq 1, \,\, \lambda^{(m)} \geq 0, \\
    & m=1,...,N_\mathrm{sv},
\end{split}
\end{equation}
where $x^{\mathrm{sv},(m)}$ is a state of the m-th surrounding vehicle, $A^{(m)}$ and $b^{(m)}(\cdot)$ are parameters that define the enlarged polytope of the m-th surrounding vehicle, $\lambda^{(m)}$ is a dual variable that corresponds to the m-th surrounding vehicle, and $N_\mathrm{sv}$ is the number of the surrounding vehicles.
To sum up, the feasible set can be defined as: $\mathcal{F}^\mathrm{LC}(\{x^{\mathrm{sv},(m)}\}_{m=1}^{N_\mathrm{sv}}) =\{(\mathbf{x}, \mathbf{u}): \eqref{eq: LK state constraint}, \eqref{eq: LK input constraint}, \eqref{eq: LC state input ay constraint}, \eqref{eq:obca constraint}\}$

The terminal state should be in the target lane with small path tracking errors as follows: 
\begin{equation} \label{eq: LC lateral terminal set}
    \begin{split}
        & |e_{y, N|k} - y_\mathrm{target}| \leq 0.1, \,\, |e_{\psi, N|k}| \leq 0.1
    \end{split}
\end{equation}
Moreover, the ego vehicle should be in a space that is predicted to be free.
If there is at least one surrounding vehicle, then the number of free spaces is at least two.
Choosing appropriate free spaces can be included in the optimization problem but this formulation involves integer programming.
Thus, we use the heuristic rules in \cite{spaceselection}, which are based on analysis of the human driving data, to select the free space.
Given the free space, a constraint is imposed on the terminal traveled distance $s_{N|k}$ as:
\begin{equation} \label{eq: LC longitudinal terminal set}
    \begin{split}
        & s_\mathrm{min}^\mathrm{free} \leq s_{N|k} \leq s_\mathrm{max}^\mathrm{free},
    \end{split}
\end{equation}
where $s_\mathrm{max}^\mathrm{free}$ and $s_\mathrm{min}^\mathrm{free}$ are the traveled distance values of boundaries of the selected free space.
To sum up, the terminal set can be defined as: $\mathcal{X}^\mathrm{LC}_f =\{\mathbf{x}: \eqref{eq: LC lateral terminal set}, \eqref{eq: LC longitudinal terminal set}\}$.

\subsubsection{Optimal Control Problem}
To sum up, we solve the following constrained optimal control problem to plan a lane change trajectory:
\begin{equation} \label{eq:lane change planner}
\begin{split}
    & V^\mathrm{LC}_{k \rightarrow k+N} (\mathbf{x}_k, \mathbf{u}_k) = \\
    & \min_{\{\mathbf{u}_{i|k}\}_{i=0}^{N-1}, \{\lambda\}} \,\, J^\mathrm{LC}(\{\mathbf{x}_{i|k}\}_{i=0}^{N}, \{\mathbf{u}_{i|k}\}_{i=0}^{N-1}) \\
    & \qquad \textnormal{s.t.,} \,\,\,  \mathbf{x}_{0|k} = \mathbf{x}_k, \,\, \mathbf{u}_{0|k} = \mathbf{u}_k,\\
    & \qquad \qquad \mathbf{x}_{i+1|k}  = \mathbf{f}_d(\mathbf{x}_{i|k}) + \mathbf{g}_d(\mathbf{x}_{i|k}) \mathbf{u}_{i|k},\\
    & \qquad \qquad (\mathbf{x}_{i|k}, \mathbf{u}_{i|k}) \in \mathcal{F}^\mathrm{LC}(\{x^{\mathrm{sv},(m)}\}_{m=1}^{N_\mathrm{sv}}),  \\
    & \qquad \qquad \mathbf{x}_{N|k} \in \mathcal{X}^\mathrm{LC}_f,  \\
    & \qquad \qquad  i=0,1,...,N-1, \,\, m=1,...,N_\mathrm{sv} 
\end{split}
\end{equation}
After solving \eqref{eq:lane change planner}, the planner transmits whole optimal sequences of state $\{\mathbf{x}^{\star}_{i|k}\}_{i=0}^{N}$ and input $\{\mathbf{u}^{\star}_{i|k}\}_{i=0}^{N-1}$ to the tracking controller.
In this paper, we do not solve \eqref{eq:lane change planner} at each time step in receding horizon manner, but solve only once and use it as a reference until the lane change maneuver is completed.
Solving the problem in receding horizon fashion also works but in practice, solving it once with conservative collision avoidance constraint (ex: set large $d_\mathrm{min}$ in \eqref{eq:obca constraint}) results in smoother behavior.
\begin{remark}
    Solving \eqref{eq:lane change planner} can fail. For example, there does not exist feasible lane change trajectory due to dense traffic, or nonlinear optimization fails to find the solution.
    In such case, we use the lane keeping problem in \eqref{eq:lane keeping planner} as a backup planner.
\end{remark}

 \subsection{Implementation}
 Both optimization problems are implemented with CasADi \cite{casadi}, and IPOPT \cite{ipopt} is used as the numerical solver.
The problems are solved in a computer with an Intel i7-9700 processor clocked up to 3.6 GHz.

\section{Experiments}
\subsection{Hardware Setup}
We build a Vehicle-In-the-Loop (VIL) system to safely and efficiently demonstrate hardware experiments under various scenarios. The system requires an actual vehicle (or multiple actual vehicles) that can autonomously drive on a specified road taking into account constraints such as traffic regulations, road geometry, and the predicted behavior of other road participants. We use Hyundai Ioniq Plug-In Hybrid as the actual vehicle. Additionally, the system utilizes microscopic simulators to generate and control virtual environments, such as other surrounding vehicles and traffic lights, in a digital twin of the real-world map. The actual vehicle and all virtual environments interact with each other in real time. 

The computing unit of the system consists of three computers: a Linux-based laptop, a Linux-based rugged computer, and the dSPACE MicroAutoBox II (MABXII).
The laptop is for simulating virtual environments and transmitting all information such as states of surrounding vehicles, SPaT messages of relevant traffic lights, and so on.
The rugged PC is for implementing a planning and control software stack that plans the ego vehicle's behavior, generates dynamically feasible, safe trajectories, and calculates acceleration and yaw rate to track the generated trajectories.
The MABXII is for implementing an actuator-level controller that calculates actuator control inputs and a fail-safe logic that provides safety features. 

The sensors of the system are an OxTS RT3000: a differential GPS to localize the ego vehicle, a Cohda MK5: a Vehicle-to-Infrastructure (V2I) communication module using Dedicated Short-Range Communications (DSRC) technology to receive SPaT messages of actual traffic lights for synchronizing the actual and simulated traffic light, and production vehicle sensors to acquire vehicle state information. 


\subsection{Virtual Environment Simulator (Digital Twin)}
\begin{figure}[h]
    \centering    \includegraphics[width=0.80\columnwidth]{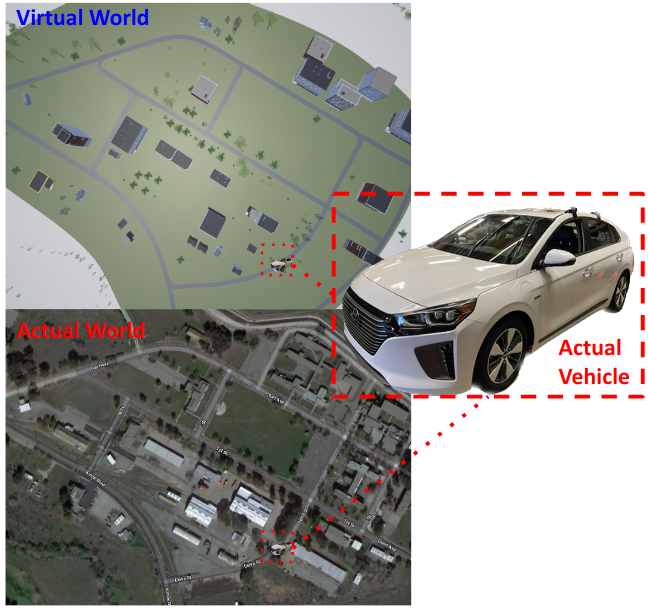}
    \caption{The CARLA image, the satellite image of the testing site and the actual vehicle image}
    \label{fig:compare_map}
\end{figure}

The CARLA software is the primary simulator to build virtual environments and simulate a variety of scenarios with ease. 
The virtual environment simulator constructs all components such as road networks, other vehicles, traffic infrastructures, buildings, and so on to replicate the real-world map. 
Fig. \ref{fig:compare_map} shows the generated CARLA map, the satellite image of the testing site, and the image of the actual test vehicle, the Hyundai Ioniq Plug-In Hybrid.

On the customized map, the CARLA simulates a traffic scenario with the same initial condition such as the number of spawned vehicles, the locations of the vehicles, the cycle period of traffic lights, etc.
It is worth noting that the CARLA simulator exhibits inherent randomness in the motion of each virtual vehicle, resulting in variations in the resulting traffic scenario. 
We also synchronize the real world with the virtual world in terms of the physical ego vehicle and the traffic lights.
Based on the obtained coordinate data from dGPS/IMU sensors, the simulator generates an agent in the virtual world and teleports the vehicle by updating the position and orientation of the agent every time it receives data from the actual sensors.

In this way, our system is able to effectively mimic the behavior and reactions of the ego vehicle in response to virtual traffic and virtual traffic lights, making it possible to test and evaluate autonomous energy-efficient driving algorithms in a safe and controlled environment.

\subsection{Results}
We evaluated the proposed algorithm using the VIL setup with the test vehicle shown in Fig. \ref{fig:compare_map}.
The proposed algorithm has been compared with a baseline algorithm, which is a lane-keeping algorithm that generates a trajectory by solving \eqref{eq:lane keeping planner} without considering the energy stage cost \eqref{eq: lk energy cost}.
We place the test vehicle in the same position and let it autonomously drive a 4km route at the testing site, repeated four times each.
The traffic is generated identically for each scenario, but the resulting traffic of each test could be different as each virtual vehicle chooses its motion randomly.

\begin{figure}[h]
    \centering    
    \includegraphics[width=1.0\columnwidth]{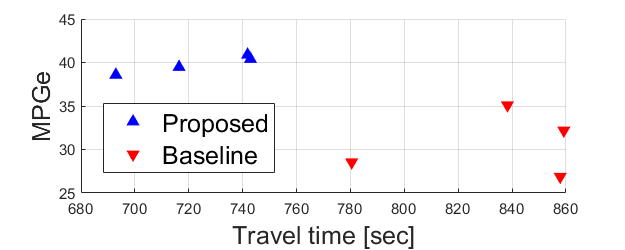}
    \caption{Overall MPGe Comparison between Baseline and Proposed}
    \label{fig:compare_mpge}
\end{figure}

\begin{figure}[h]
    \centering    
    \includegraphics[width=1 \columnwidth]{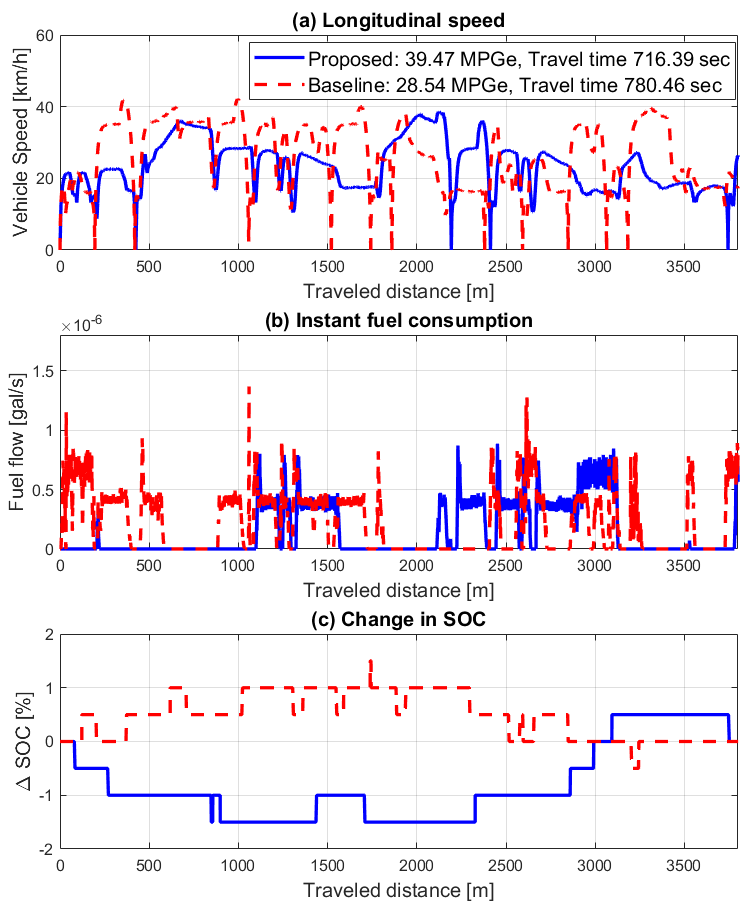}
    \caption{Analysis: Changing lanes to minimize the number of stops is energy-efficient.}
    \label{fig:analyze energy saving}
\end{figure}

The test results are shown in Fig. \ref{fig:compare_mpge} and Fig. \ref{fig:analyze energy saving}. 
Fig. \ref{fig:compare_mpge} presents a Mile Per Gallon equivalent (MPGe) result for each test.
In Fig. \ref{fig:analyze energy saving}, we pick one test case that corresponds to the median MPGe for each algorithm and analyzes the results.

As shown in Fig. \ref{fig:compare_mpge}, the proposed algorithm shows better MPGe than the baseline.
The average MPGe of the proposed algorithm is 39.81 MPGe (37.1\% improvement), while that of the baseline is 29.04 MPGe.

As demonstrated in the human driving data in Fig. \ref{fig:human-driving}, reducing the number of stops is important to reduce energy consumption.
As shown in Fig. \ref{fig:analyze energy saving}, the proposed algorithm well minimizes the number of stops by changing lanes timely: the number of the stops of the proposed algorithm is 5 while that of the baseline is 13.
Moreover, as illustrated in Fig. \ref{fig:analyze energy saving} (b), stops result in a loss of kinetic energy, requiring additional energy consumption to accelerate again.
For example, as shown in the traveled distance between $1600m$ and $1800m$ of Fig. \ref{fig:analyze energy saving}, the baseline algorithm spent considerable fuel to accelerate after a full stop, while the proposed algorithm did not have to. As an additional benefit of minimizing the number of stops, the average traveled time of the proposed algorithm (12min 3sec) is 13\% less than that of the baseline (13min 53sec).

\section{Conclusion and Future work}
The novel energy-efficient motion planning algorithm for CAVs is presented.
The proposed algorithm consists of the lane selector, which selects a lane to reduce energy consumption by minimizing the number of stops, and the energy-efficient trajectory planner, which generates a safe, smooth, and energy-efficient trajectory toward the selected lane.
The proposed algorithm is experimentally evaluated in the VIL setting.
Compared to a lane-keeping algorithm, an average 37.1\% of energy savings is measured with the proposed algorithm.
As the energy-efficient motion planning algorithm using SPaT information for CAVs has been successfully developed, our future works are targeted to using the full capability of the connectivity including V2V and V2C communications.
Additionally, further research will aim to establish a systematic method for lane selection, beyond the current heuristic approach, and design the terminal cost and the terminal set for approximating energy consumption in long horizons. 
\section*{ACKNOWLEDGMENT}
This research work presented herein is funded
by the Advanced Research Projects Agency-Energy 
(ARPA-E), U.S. Department of Energy under DE-AR0000791. 
\bibliography{IEEEabrv,reference.bib}

\begin{thebibliography}{10}
\providecommand{\url}[1]{#1}
\csname url@samestyle\endcsname
\providecommand{\newblock}{\relax}
\providecommand{\bibinfo}[2]{#2}
\providecommand{\BIBentrySTDinterwordspacing}{\spaceskip=0pt\relax}
\providecommand{\BIBentryALTinterwordstretchfactor}{4}
\providecommand{\BIBentryALTinterwordspacing}{\spaceskip=\fontdimen2\font plus
\BIBentryALTinterwordstretchfactor\fontdimen3\font minus
  \fontdimen4\font\relax}
\providecommand{\BIBforeignlanguage}[2]{{%
\expandafter\ifx\csname l@#1\endcsname\relax
\typeout{** WARNING: IEEEtran.bst: No hyphenation pattern has been}%
\typeout{** loaded for the language `#1'. Using the pattern for}%
\typeout{** the default language instead.}%
\else
\language=\csname l@#1\endcsname
\fi
#2}}
\providecommand{\BIBdecl}{\relax}
\BIBdecl

\bibitem{guanetti2018CAVs}
\BIBentryALTinterwordspacing
J.~Guanetti, Y.~Kim, and F.~Borrelli, ``Control of connected and automated
  vehicles: State of the art and future challenges,'' \emph{Annual Reviews in
  Control}, vol.~45, pp. 18--40, 2018. [Online]. Available:
  \url{https://www.sciencedirect.com/science/article/pii/S1367578818300336}
\BIBentrySTDinterwordspacing

\bibitem{naus2010string}
G.~J. Naus, R.~P. Vugts, J.~Ploeg, M.~J. van De~Molengraft, and M.~Steinbuch,
  ``String-stable cacc design and experimental validation: A frequency-domain
  approach,'' \emph{IEEE Transactions on vehicular technology}, vol.~59, no.~9,
  pp. 4268--4279, 2010.

\bibitem{mcauliffe2018CACC}
B.~McAuliffe, M.~Lammert, X.-Y. Lu, S.~Shladover, M.-D. Surcel, and A.~Kailas,
  ``Influences on energy savings of heavy trucks using cooperative adaptive
  cruise control,'' \emph{SAE technical paper}, no. 2018-01, p. 1181, 2018.

\bibitem{kim2021CACC}
Y.~Kim, J.~Guanetti, and F.~Borrelli, ``Compact cooperative adaptive cruise
  control for energy saving: Air drag modelling and simulation,'' \emph{IEEE
  Transactions on Vehicular Technology}, vol.~70, no.~10, pp. 9838--9848, 2021.

\bibitem{askari2017CACCthroughput}
A.~Askari, D.~A. Farias, A.~A. Kurzhanskiy, and P.~Varaiya, ``Effect of
  adaptive and cooperative adaptive cruise control on throughput of signalized
  arterials,'' in \emph{2017 IEEE Intelligent Vehicles Symposium (IV)}, 2017,
  pp. 1287--1292.

\bibitem{bertoni2017CACCurban}
L.~Bertoni, J.~Guanetti, M.~Basso, M.~Masoero, S.~Cetinkunt, and F.~Borrelli,
  ``An adaptive cruise control for connected energy-saving electric vehicles,''
  \emph{IFAC-PapersOnLine}, vol.~50, no.~1, pp. 2359--2364, 2017.

\bibitem{sciarretta2020CAVs}
A.~Sciarretta and A.~Vahidi, \emph{Energy-Efficient Driving of Road Vehicles:
  Toward Cooperative, Connected, and Automated Mobility}.\hskip 1em plus 0.5em
  minus 0.4em\relax Springer International Publishing, 2020.

\bibitem{bae2019real}
S.~Bae, Y.~Choi, Y.~Kim, J.~Guanetti, F.~Borrelli, and S.~Moura, ``Real-time
  ecological velocity planning for plug-in hybrid vehicles with partial
  communication to traffic lights,'' in \emph{2019 IEEE 58th Conference on
  Decision and Control (CDC)}.\hskip 1em plus 0.5em minus 0.4em\relax IEEE,
  2019, pp. 1279--1285.

\bibitem{ard2023VILCAV}
T.~Ard, L.~Guo, J.~Han, Y.~Jia, A.~Vahidi, and D.~Karbowski, ``Energy-efficient
  driving in connected corridors via minimum principle control:
  Vehicle-in-the-loop experimental verification in mixed fleets,'' \emph{IEEE
  Transactions on Intelligent Vehicles}, pp. 1--14, 2023.

\bibitem{wang2019V2IV2V}
Z.~Wang, G.~Wu, and M.~J. Barth, ``Cooperative eco-driving at signalized
  intersections in a partially connected and automated vehicle environment,''
  \emph{IEEE Transactions on Intelligent Transportation Systems}, vol.~21,
  no.~5, pp. 2029--2038, 2019.

\bibitem{ma2018VIL}
J.~Ma, F.~Zhou, Z.~Huang, C.~L. Melson, R.~James, and X.~Zhang,
  ``Hardware-in-the-loop testing of connected and automated vehicle
  applications: a use case for queue-aware signalized intersection approach and
  departure,'' \emph{Transportation Research Record}, vol. 2672, no.~22, pp.
  36--46, 2018.

\bibitem{bae2019VIL}
S.~Bae, Y.~Kim, J.~Guanetti, F.~Borrelli, and S.~Moura, ``Design and
  implementation of ecological adaptive cruise control for autonomous driving
  with communication to traffic lights,'' in \emph{2019 American Control
  Conference (ACC)}, 2019, pp. 4628--4634.

\bibitem{ard2021energyVIL}
T.~Ard, L.~Guo, R.~A. Dollar, A.~Fayazi, N.~Goulet, Y.~Jia, B.~Ayalew, and
  A.~Vahidi, ``Energy and flow effects of optimal automated driving in mixed
  traffic: Vehicle-in-the-loop experimental results,'' \emph{Transportation
  Research Part C: Emerging Technologies}, vol. 130, p. 103168, 2021.

\bibitem{bae2022ecological}
S.~Bae, Y.~Kim, Y.~Choi, J.~Guanetti, P.~Gill, F.~Borrelli, and S.~J. Moura,
  ``Ecological adaptive cruise control of plug-in hybrid electric vehicle with
  connected infrastructure and on-road experiments,'' \emph{Journal of Dynamic
  Systems, Measurement, and Control}, vol. 144, no.~1, p. 011109, 2022.

\bibitem{kim2023trajectory}
C.~Kim, Y.~Yoon, S.~Kim, M.~J. Yoo, and K.~Yi, ``Trajectory planning and
  control of autonomous vehicles for static vehicle avoidance in dynamic
  traffic environments,'' \emph{IEEE Access}, 2023.

\bibitem{adaptive_control_steering_control}
E.~Joa, K.~Yi, and K.~Kim, ``A lateral driver model for vehicle--driver
  closed-loop simulation at the limits of handling,'' \emph{Vehicle system
  dynamics}, vol.~53, no.~9, pp. 1247--1268, 2015.

\bibitem{gonzalez2014continuous}
D.~Gonz{\'a}lez, J.~P{\'e}rez, R.~Lattarulo, V.~Milan{\'e}s, and F.~Nashashibi,
  ``Continuous curvature planning with obstacle avoidance capabilities in urban
  scenarios,'' in \emph{17th International IEEE Conference on Intelligent
  Transportation Systems (ITSC)}.\hskip 1em plus 0.5em minus 0.4em\relax IEEE,
  2014, pp. 1430--1435.

\bibitem{sciarretta2015optimal}
A.~Sciarretta, G.~De~Nunzio, and L.~L. Ojeda, ``Optimal ecodriving control:
  Energy-efficient driving of road vehicles as an optimal control problem,''
  \emph{IEEE control systems magazine}, vol.~35, no.~5, pp. 71--90, 2015.

\bibitem{Choi2021Thesis}
Y.~Choi, ``Energy efficient vehicle dynamics and powertrain controls for
  connected plug-in hybrid electric vehicles,'' Ph.D. dissertation, UC
  Berkeley, 2021.

\bibitem{de2023standards}
K.~N. de~Winkel, T.~Irmak, R.~Happee, and B.~Shyrokau, ``Standards for
  passenger comfort in automated vehicles: Acceleration and jerk,''
  \emph{Applied Ergonomics}, vol. 106, p. 103881, 2023.

\bibitem{obca}
X.~Zhang, A.~Liniger, and F.~Borrelli, ``Optimization-based collision
  avoidance,'' \emph{IEEE Transactions on Control Systems Technology}, vol.~29,
  no.~3, pp. 972--983, 2020.

\bibitem{spaceselection}
H.~Chae, Y.~Jeong, H.~Lee, J.~Park, and K.~Yi, ``Design and implementation of
  human driving data--based active lane change control for autonomous
  vehicles,'' \emph{Proceedings of the Institution of Mechanical Engineers,
  Part D: Journal of Automobile Engineering}, vol. 235, no.~1, pp. 55--77,
  2021.

\bibitem{casadi}
J.~A.~E. Andersson, J.~Gillis, G.~Horn, J.~B. Rawlings, and M.~Diehl,
  ``{CasADi} -- {A} software framework for nonlinear optimization and optimal
  control,'' \emph{Mathematical Programming Computation}, In Press, 2018.

\bibitem{ipopt}
A.~W{\"a}chter and L.~T. Biegler, ``On the implementation of an interior-point
  filter line-search algorithm for large-scale nonlinear programming,''
  \emph{Mathematical programming}, vol. 106, no.~1, pp. 25--57, 2006.

\end{thebibliography}
\end{document}